\documentclass{article}

\PassOptionsToPackage{numbers, compress}{natbib}
\usepackage[utf8]{inputenc} % allow utf-8 input
\usepackage[T1]{fontenc}    % use 8-bit T1 fonts

\usepackage{hyperref}       % hyperlinks
\usepackage{url}            % simple URL typesetting
\usepackage{booktabs}       % professional-quality tables
\usepackage{amsfonts}       % blackboard math symbols
\usepackage{nicefrac}       % compact symbols for 1/2, etc.
\usepackage{microtype}      % microtypography
\usepackage{xcolor}         % colors
\usepackage[preprint]{neurips_2022}
 
\usepackage{microtype}
\usepackage{bbding}
\usepackage{subfigure}
\usepackage{graphicx}
\usepackage{amsmath}
\usepackage{bm}

\newcommand{\ten}[1]{\bm{\mathcal{#1}}}
\usepackage{amssymb}
\usepackage{booktabs}
\usepackage{multirow}
\usepackage{upgreek}
\usepackage{mathtools}
\usepackage{lipsum}
\usepackage{algorithm}
\usepackage{algorithmic}
\usepackage{xspace}
\usepackage{wrapfig}
\title{PECAN: A Product-Quantized Content Addressable Memory Network}

\author{%
Jie Ran, \quad Rui Lin,  \quad Jason Chun Lok Li, \quad Jiajun Zhou, \quad Ngai Wong \\
Department of Electrical and Electronic Engineering, \\
The University of Hong Kong, Hong Kong \\
Email Address: \texttt{\{jieran, linrui,  u3524157, jjzhou, nwong\}@eee.hku.hk}\\}

\begin{document}
\maketitle

%%%%%%%%% ABSTRACT %%%%%%%%%
\begin{abstract}
A novel deep neural network (DNN) architecture is proposed wherein the filtering and linear transform are realized solely with product quantization (PQ). This results in a natural implementation via content addressable memory (CAM), which transcends regular DNN layer operations and requires only simple table lookup. Two schemes are developed for the end-to-end PQ prototype training, namely, through angle- and distance-based similarities, which differ in their multiplicative and additive natures with different complexity-accuracy tradeoffs. Even more, the distance-based scheme constitutes a truly multiplier-free DNN solution. Experiments confirm the feasibility of such \textbf{P}roduct-Quantiz\textbf{E}d \textbf{C}ontent \textbf{A}ddressable Memory \textbf{N}etwork (PECAN), which has strong implication on hardware-efficient deployments especially for in-memory computing.
\end{abstract}

% %%% comment this command when submitting
% \pagestyle{myheadings}

%%%%%%%%% BODY TEXT

%%%%% Introduction %%%%% 
\section{Introduction}
\label{sec:intro}
Deep neural networks (DNNs) have achieved breakthroughs in various applications including classification~\cite{sun2020automatically}, object detection~\cite{liu2020cbnet} and semantic segmentation~\cite{zheng2021rethinking}, etc. Nonetheless, the massive amount of parameters and computation make it difficult for both training and inference on edge devices with constrained hardware resources. Numerous efforts have been made to reduce the network complexity while preserving the output accuracy. Among various schemes, some are low-bitwidth neural networks using binary weights~\cite{courbariaux2015binaryconnect, rastegari2016xnor, xue2021self}, replacing the expensive multiplications with cheaper sign flip operations during inference. Some approaches substitute multiplications with additions and bit-wise shifts. AdderNet~\cite{chen2020addernet} realizes convolution (in the sense of similarity matching) by $l_1$-distance between the activation and weights, and maintains competitive output accuracy. ShiftCNN~\cite{gudovskiy2017shiftcnn} is based on a power-of-two weight representation for converting convolutional neural networks (CNNs) without retraining. Among works that aim to improve the memory efficiency and performance of shift neural networks,  DeepShift~\cite{elhoushi2021deepshift} is a framework for training low-bitwidth neural networks from scratch to replace multiplication with bit-wise shift and sign flip. All these works, despite specific implementations, still adhere to the traditional DNN architecture.
% ShiftAddNet~\cite{you2020shiftaddnet} combines bit-shift-based and add-based networks to implement the shift and add layers.   

This work attempts to detach a neural network from its regular filtering operation and replace it with an associative memory, aka content addressable memory (CAM), whereby the content is derived from prototypes of product quantization~\cite{jegou2010product}. Such framework, dubbed \textbf{P}roduct-Quantiz\textbf{E}d \textbf{C}ontent \textbf{A}ddressable Memory \textbf{N}etwork (PECAN), combines the storage and compute into one place, and is particularly suitable for the fast-emerging in-memory computing. The codebook/table lookup during inference also makes PECAN hardware-friendly and positions it as a strong candidate for edge artificial intelligence (AI). This is also warranted by the readiness in commodity platforms like FPGAs with CAM support, as well as next-generation memristive microelectronics like resistive random-access memory (RRAM) wherein a CAM is inherent to an RRAM crossbar~\cite{batmann, Karunaratne2021RobustHM}.

Our proposed PECAN is inspired by the lately proposed MADDNESS~\cite{blalock2021multiplying} that utilizes product quantization and table lookup to truly omit multipliers in matrix-matrix products. However, the main contribution of MADDNESS, namely, the hash function for prototype matching, is heuristic and non-differentiable, thus making it incompatible with a learning framework. In fact, the authors also remark it will take several more papers to consolidate the framework for DNNs.

PECAN exactly fills this void by its end-to-end learnable PQ-based DNN architecture. The closest work to ours is differentiable product quantization (DPQ)~\cite{chen2020differentiable}, but \emph{for the first time} we demonstrate its multi-layer feasibility and enrich DPQ prototype matching (viz. a similarity search) with an $l_1$-distance metric. The latter comes from the lately proposed AdderNet~\cite{chen2020addernet} wherein the $l_1$ metric is utilized in a different context of CNN filtering, whereas our work is the first to show its feasibility for training prototypes in the DPQ setting. To our best knowledge, PECAN is a brand new architecture that transcends regular DNN filtering and uses similarity search and table lookup for inference. This allows it to be compatible with simple hardware without the need of dedicated neural engines, especially edge devices where compute and storage resources are limited. Our major contributions are: 1) A first-of-its-kind, end-to-end learnable CAM-based DNN. PECAN is hardware-generic and friendly to almost all hardware platforms especially those with built-in CAM support, and represents a strong candidate for edge AI deployment; 2) Two similarity measures in PECAN, based on angle and distance, to investigate the trade-offs between computation complexity and accuracy; 3) Joint fine-tuning and co-optimization of weight matrices and PQ prototypes, which permits PECAN to train from scratch; 4) A \emph{totally} multiplier-free DNN via the distance-based PECAN.

%In short, PECAN represents a brand new DNN architecture that decouples itself from regular filtering and transform, and replaces them by querying pretrained codebooks.
%It is hopeful that further development and variants of PECAN will come along when more attention is drawn to it.

%%%%% Related Work %%%%% 
\section{Related Work}
\label{sec:lit_review}
For efficient edge deployment, binary neural networks (BNNs)~\cite{hubara2016binarized, rastegari2016xnor, qin2020forward} exclusively make use of the logical XNOR operation that obviates regular multipliers, but in principle they are still doing 1-bit multiplication. Moreover, though BNNs have gone through major improvements in recent years, their top-1 accuracies measured on large-scale datasets are still noticeably lower than their full-precision counterparts. Indeed, most BNN implementations are only partial in the sense that the first and final layers are still using full-precision weights and activations~\cite{xue2021self, xu2021learning}.

Other works replace multiplication with addition~\cite{chen2020addernet} or bit-shift operations~\cite{gudovskiy2017shiftcnn, zhou2017incremental, elhoushi2021deepshift}, or both~\cite{you2020shiftaddnet}. Specifically, AdderNet makes novel use of $l_1$-norm difference and adders to do template matching required in a CNN. Yet it still employs multipliers for the necessary batch normalization to bring back signed pre-activations. Progressive kernel based knowledge distillation (PKKD)  AdderNet~\cite{xu2020kernel} improves the performance of the vanilla AdderNet. AdderNet with Adaptive Weight Normalization (AWN)~\cite{dong2021towards} further alleviates the curse of instability of running mean and variance in batch normalization layers. Applying bitwise shift on an element is mathematically equivalent to multiplying it by a power of two, and sign flipping is introduced to represent negative numbers. Although these works focus on largely multiplier-free DNNs, they still build on the traditional architectures.
% while our distance-based measure requires only lookup table and addition\textcolor{red}{<-- strange sentence}. 

The proposed PECAN is motivated by MADDNESS which realizes multiplier-free matrix-matrix product using hashing and table lookup rather than multiply-add operations. Although it achieves orders of speedups compared to existing approximate matrix multiplication (AMM) methods, the proposed hashing functions are not differentiable and not amenable to DNN training. DPQ~\cite{chen2020differentiable} is proposed for end-to-end embedding, but it is only single-layer and targets word embedding, and still requires full-precision multiplication to obtain distances between the input and matching keys.

\begin{figure}[t]
    \centering
    \includegraphics[scale=0.5]{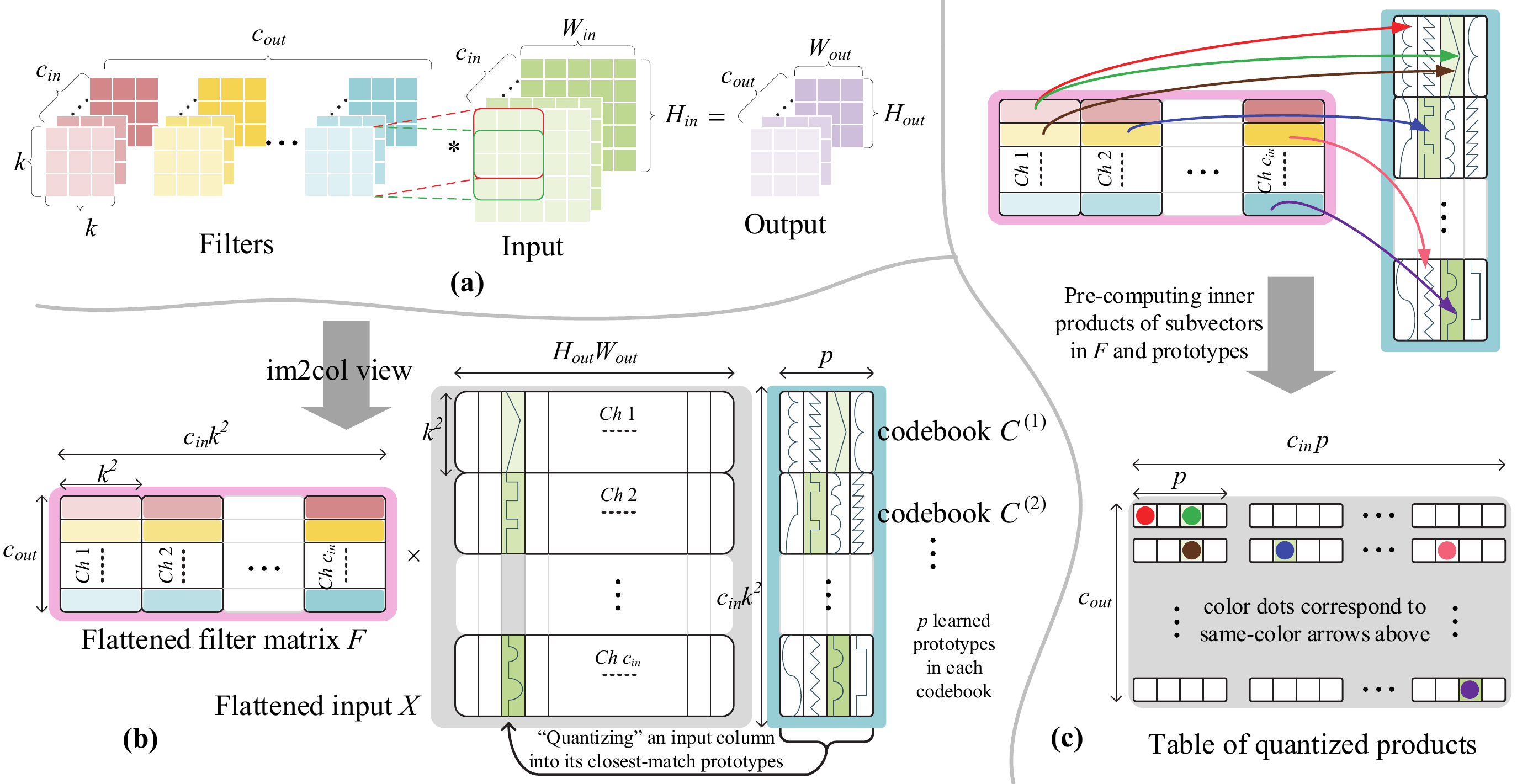}
    \vspace{-2mm}
    \caption{CNN convolution in its (a) conceptual form; (b) equivalent matrix-matrix-multiply by flattening the filters and input features via {\tt im2col}, with mapping of input data sub-columns onto the closest prototypes in different codebooks; (c) Precomputing inner products of $F$-subvectors and prototypes to form a lookup table.}
    \label{fig:gemm}
    \vspace{-2mm}
\end{figure}
%%%%% Methodology %%%%% 
\section{PECAN}
\label{sec:PECAN}
% \textcolor{red}{This section should begin with a description of CNN and its im2col equivalent so that reader gets the picture of matrix-matrix multiplication. Can use a diagram like in DeBut (perhaps Rui redraws that?) for illustration. }
%the matrix-matrix product required in a CNN layer by transforming the input tensor into an {\tt im2col}-flattened matrix $X$ while
The convolution operation in a CNN is conceptually illustrated as a window sliding across the $c_{in}$-channel input feature (cf. Fig.~\ref{fig:gemm}(a)). Actual implementations often unfold the convolution into a matrix-matrix product (cf. Fig.~\ref{fig:gemm}(b)). Specifically, the {\tt im2col} command stretches the input entries covered in each filter stride into a column and concatenates the columns into a matrix $X$, whereas the kernel tensors are reshaped into a filter matrix $F$, such that PQ can be used to approximate $FX$. For an intermediate CNN layer, consider the flattened feature matrix $X \in \mathbb{R}^{c_{in} k^2 \times H_{out} W_{out}}$, where $c_{in}$ and $k$ are the number of input channels and the kernel size, $H_{out}$ and $W_{out}$ are height and width of the output feature, codebooks $C \in \mathbb{R}^{c_{in} k^2 \times p}$ are assigned with parameters to construct an embedding table for the features, where $p$ is the number of choices for each \textit{codebook} $C^{(j)}$, $j = 1, 2, \ldots, D$. $C_m^{(j)} \in \mathbb{R}^d$ are called \textit{prototypes}, $m = 1,2,\ldots, p$ (cf. Fig.~\ref{fig:gemm}(c)). It is natural to set each prototype in PQ to be a $k^2\times 1$ subvector (viz. same size as a vectorized kernel), with $p$ prototypes in each of the $c_{in}$ input channels according to the patterns of flattened matrices. With this setting, there are two main components in a trained PECAN that require memory storage in each layer, namely, i) $pc_{in}$ prototypes for ``quantizing'' the input subvectors; ii) $c_{out}c_{in}p$ inner product values between the (sub)rows in $F$ and each prototype.

In short, PECAN is mapping (quantizing) the original input features onto prototype patterns in compact codebooks, then multiplication between weights ($F$) and features ($X$) can be approximated by lookup table operation during inference. Below we elaborate two content addressing techniques (i.e. similarity matching) approaches based respectively on angle (dot product) and distance ($l_1$-norm) which are both end-to-end learnable. Accordingly, these two schemes are dubbed PECAN-A and PECAN-D, which cover both ends of complexity-accuracy spectrum: The angle-based scheme uses multiplicative operations and generally leads to higher output accuracy, whereas the distance-based one uses additive operations and is much more lightweight at the expense of slight accuracy loss. 

% \begin{figure}[t]
% \centering
%   \includegraphics[scale=.43]{figures/CoBoNet.png}
%   \caption{The mapping of input data columns onto prototype subvectors in a CNN layer under the {\tt im2col} setting.}
%   \label{fig:cnnLayer}
% \end{figure}

%% workflow placeholder
\begin{figure*}[t]
\centering
\includegraphics[scale=0.8]{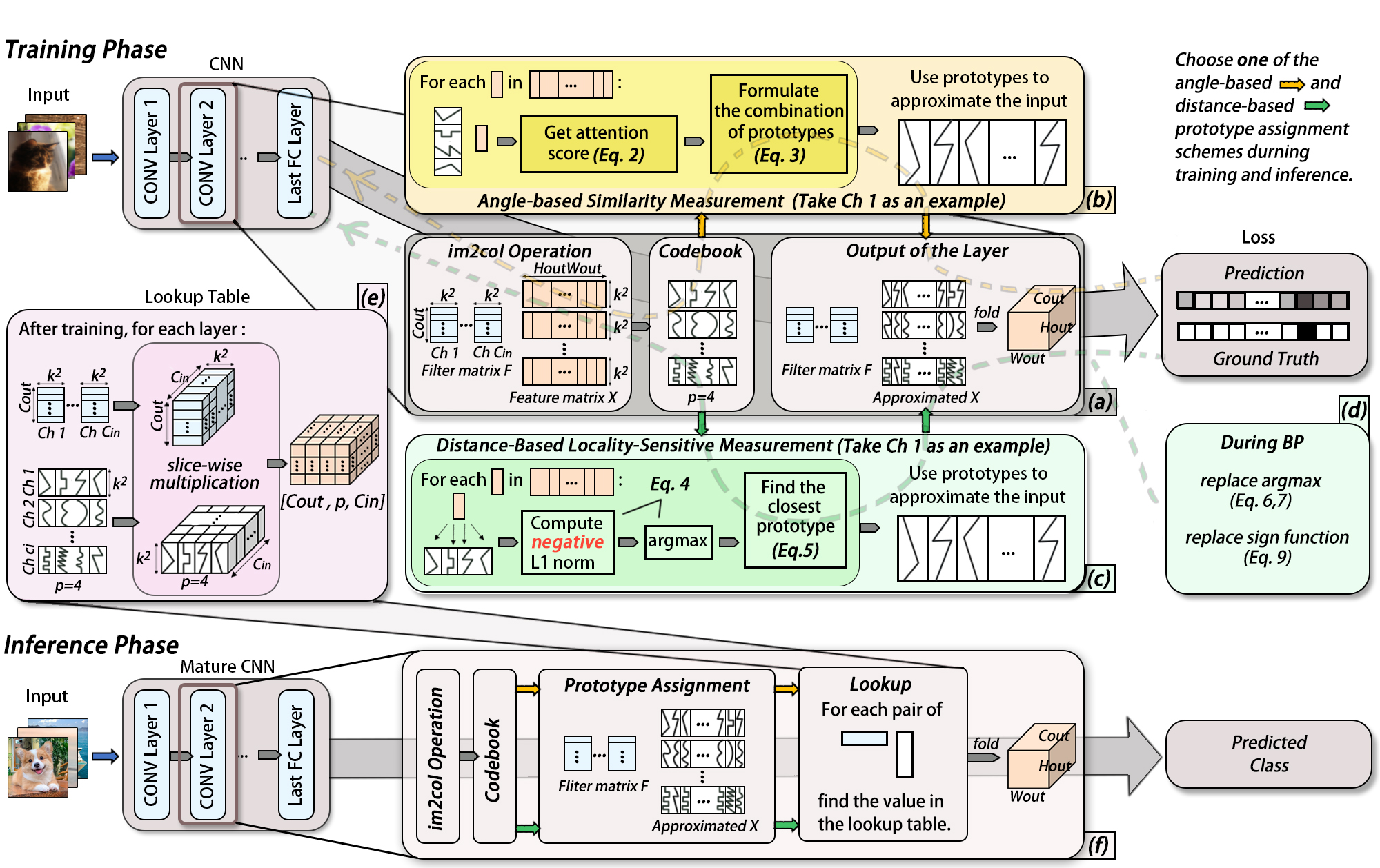}
\caption{The proposed PECAN architecture. \textbf{(a)} The training phase is mainly composed of template matching for each subvector in the flattened feature map matrices after {\tt im2col} operation. When approximating subvectors with the closest prototypes, PECAN-A and PECAN-D adopt different assignment schemes. \textbf{(b)} For PECAN-A, an attention module compares the subvectors with each of the prototypes in the same group. Subsequently, the resulting scores are subjected to the weighted sum to produce the approximate feature matrix. \textbf{(c)} For PECAN-D, the similarity is measured with a sign flip $l_1$-norm and the approximation is selected with {\tt argmax} function. \textbf{(d)} Since the {\tt argmax} is not differentiable and the gradient of $l_1$-norm is discrete ($1, -1, 0$), we propose Eq.~(\ref{eq:dif_softmax},~\ref{eq:dif_softmax2}) and (\ref{eq:tanh}) to do the backpropagation. \textbf{(e)} After getting the converged neural network, we calculate the slice-wise product between convolution filters and prototypes, and store the results in the memory. \textbf{(f)} In the inference phase, we only need to calculate the distance of feature maps with a small number of prototypes and look up in the stored memory to get the quantized output.}
\vspace{-2mm}
\label{fig:workflow}
\end{figure*}

\subsection{PECAN-A: Angle-Based Similarity Measure}
A scaled dot-product attention module~\cite{vaswani2017attention}, widely used in Transformers, computes the dot products of queries with keys and applies a {\rm softmax} function to obtain the weights on the values. \begin{equation}
    {\rm Attention} (Q, K, V ) = {\rm softmax} (\frac{QK^T}{\sqrt{d_k}}) {V},%
\end{equation}
where $d_k$ is the dimension of keys, which serves as a scaling factor. Generally, $Q$, $K$ and $V$ are obtained from three distinct learned projection matrices. However, different from self-attention, we learn the keys $K$ (viz. prototypes in PQ) directly without the intermediate linear transforms, and make $V$ equal to $K$. For PECAN-A, we compute the approximated matrix $\Tilde{X}$ by splitting its rows into $D = c_{in}$ groups, each with subvectors of dimension $d=k^2$, and get the attention scores ${K}_i^{(j)}$ to formulate the combination of prototypes $C_m^{(j)}$:
\begin{align}
    {K}_i^{(j)} = {\rm softmax}( (C^{(j)})^T X_i^{(j)}),~\Tilde{X}_i^{(j)} = C^{(j)} {K}_i^{(j)},%
\end{align}
where $i = 1,2,\ldots, H_{out}W_{out}$. Since the dot product distance function with softmax is differentiable, mapping features to prototypes can be learned end-to-end. It is worth noting that all intermediate features are replaced with the combination of learned prototypes after training.

\subsection{PECAN-D: Distance-Based Similarity Measure}
Now we attempt to get rid of all multipliers. To achieve this, we make use of only $l_1$-norm difference for the so-called template matching, namely, finding the closest match through absolute difference which involves only subtraction. Specifically, in this distance-based framework, $l_1$-norm is applied in order to discard multiplication:
\begin{align}
    {k}_i^{(j)} = \mathop{\arg \max}_{m} -\|X_i^{(j)} - C_m^{(j)}\|_1,~\Tilde{X}_i^{(j)} = C^{(j)} {\rm one\_hot}({k}_i^{(j)}),%
\end{align}
where ${K}_i^{(j)}$ = {\rm one\_hot} $({k}_i^{(j)})$ denotes a $p$-dimensional vector with the ${k}_i^{(j)}$-th entry as 1 and others 0. To enable optimization for prototypes with the non-differentiable function ${\rm argmax}$, we approximate it with a differentiable softmax function defined as follows:
\begin{equation}
    \Tilde{K}_i^{(j)} = \frac{\exp (-\|X_i^{(j)} - C_m^{(j)}\|_1/\tau)}{\sum_{m'}\exp (-\|X_i^{(j)} - C_{m'}^{(j)}\|_1/\tau)},%
    \label{eq:dif_softmax}
\end{equation}
where $\tau$ is the temperature to relax the softmax function. Note that Eq. (\ref{eq:dif_softmax}) can be considered as the proportion of Laplacian kernels when $\tau \neq 0$ . It relies on the observation that the positive definite function $k(X_i^{(j)}, C_m^{(j)}) = \exp (-\|X_i^{(j)} - C_m^{(j)}\|_1/\tau)$ here defines an inner product and a lifting function $\phi$ such that the inner product $\langle \phi(X_i^{(j)}), \phi(C_m^{(j)})\rangle$ can be computed quickly using the kernel trick~\cite{rahimi2007random}.   

Now the approximated index $\Tilde{K}_i^{(j)}$ is fully differentiable when $\tau \neq 0$. However, this yields the combination of prototypes for $\Tilde{X}_i^{(j)}$ again, while we need $\tau \rightarrow 0$ to get discrete indices during the forward inference. To this end, we follow~\cite{chen2020differentiable} and define a new index to solve both non-differentiable and discrete problems in one go. Specifically, in the forward and backward passes during training, we adopt%
% \begin{equation}
% \Tilde{K}_i^{(j)} = \begin{dcases}
%     \frac{\exp (-d_{ik}^{(j)}/\tau)}{\sum_{k'}\exp (-d_{ik'}^{(j)}/\tau)}, & \tau \neq 0 \\
%     \mathop{\arg \max}_{k} -d_{ik}^{(j)},  & \tau = 0
% \end{dcases}
% \end{equation}
\begin{equation}
\Tilde{K}_i^{(j)}(\tau \neq 0) - sg\left(\Tilde{K}_i^{(j)}(\tau \neq 0) - \Tilde{K}_i^{(j)}(\tau = 0) \right),%
\label{eq:dif_softmax2}
\end{equation}
where $sg$ is \textit{stop gradient}, which takes the identity function in the forward pass and drops the gradient inside it in the backward pass. Based on this, we can now use the ${\rm argmax}$ function in the forward pass and softmax function during backpropagation. However, the partial derivative of the distance $d_{im}^{(j)} = -\|X_i^{(j)} - C_m^{(j)}\|_1$ with respect to codebook subvector $ C_m^{(j)}$ is a sign function:
\begin{align}
    \frac{ \partial d_{im}^{(j)}}{\partial C_m^{(j)}} = {\rm sgn}(X_i^{(j)} - C_m^{(j)}),~\frac{ \partial d_{im}^{(j)}}{\partial C_m^{(j)}} = \tanh\left( a (X_i^{(j)} - C_m^{(j)}) \right)~\mbox{where}~a = \exp(\frac{4e}{E}), \label{eq:tanh}
\end{align}

\begin{wrapfigure}{r}{9cm}
\centering
\includegraphics[scale=0.35]{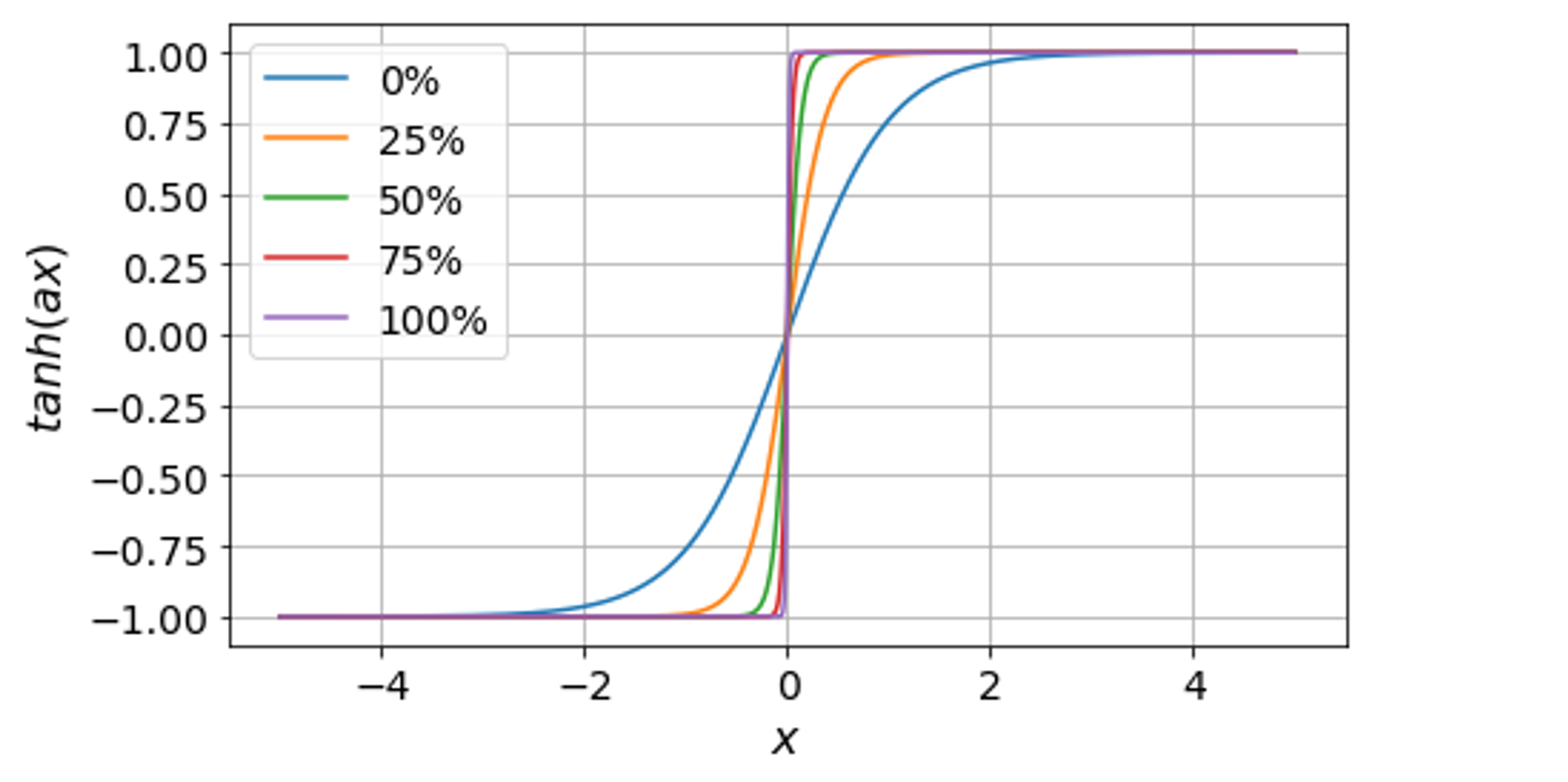}
\vspace{-3mm}
\caption{Approximation to the sign gradient for different $\frac{e}{E}$.}
\vspace{-3mm}    
\label{fig:tanh}
\end{wrapfigure}
where ${\rm sgn}(\cdot)$ is the sign function and takes the values of $\{+1, 0, -1\}$. Such zero gradient almost everywhere makes it impossible to train a neural network. In this regard, we adopt the right term in Eq.~(\ref{eq:tanh}) to replace the gradient, where $e$ is the current epoch and $E$ the total number of training epochs. Fig.~\ref{fig:tanh} shows this epoch-aware approximation to the sign function \textit{w.r.t.} values of $\frac{e}{E}$ as epoch increases during training. In the early stage, the function is smoother for stable training. As the training progresses, the approximation gradually turns into the sign-like function. 

% \begin{figure}[ht]
%     \centering
%     \includegraphics[scale=0.42]{figures/tanhx_v3.png}
%     \vspace{-5mm}
%     \caption{Approximation to the sign gradient for different $\frac{e}{E}$.}
%     \vspace{2mm}    \label{fig:tanh}
% \end{figure}
%%%%%%%%%% inference complexity %%%%%%%%%%%%
\begin{table}[t]
\caption{Inference complexities of PECAN-A and PECAN-D.}
\label{tab:inf_complexity}
\vspace{2mm}
\centering
\scriptsize
    \centering
    \setlength{\tabcolsep}{2mm}
    \renewcommand{\arraystretch}{1.4}{
    \begin{tabular}{llll}
    \toprule
        Method & Layer & \#Add. & \#Mul. \\
    \hline
        \multirow{2}{*}{Baseline} & CONV & $c_{in} H_{out} W_{out}k^2 c_{out}$ & $c_{in} H_{out} W_{out}k^2 c_{out}$ \\
        ~ & FC & $c_{in} c_{out}$ & $c_{in} c_{out}$ \\
    \hline
        \multirow{2}{*}{PECAN-A} & CONV & $p D H_{out} W_{out}(d+c_{out})$ & $p D H_{out} W_{out}(d+c_{out})$ \\
        ~ & FC & $p D (d+c_{out})$ & $p D (d+c_{out})$ \\
    \hline
        \multirow{2}{*}{PECAN-D} & CONV & $D H_{out} W_{out}(2pd + c_{out})$ & $0$ \\
        ~ & FC & $D (2pd+c_{out})$ & $0$ \\
    \bottomrule
    \end{tabular}
    }
\end{table}
\subsection{Inference Details and Complexity}
% To get rid of all multiplication in PECAN-D, we adopt batch normalization folding~\cite{BN_folding}. Specifically, given an input $x \in \mathbb{R}^{H_{in} \times W_{in} \times c_{in}}$ before a convolution layer, we have the output:
% \begin{align}
%     z &= W \ast x + b, \\
%     out &= \gamma \frac{z - \mu_{\mathcal{B}}}{\sqrt{\sigma_{\mathcal{B}}^2 + \epsilon}} + \beta,
% \end{align}
% where $W\in \mathbb{R}^{c_{out} \times c_{in} \times k \times k}$ and $b\in \mathbb{R}^{c_{out}}$ are the convolutional filter and bias, $z \in \mathbb{R}^{H_{out} \times W_{out} \times c_{out}}$ is the convolution output, $\gamma$ and $\beta$ are batch normalization parameters to be learned, $\mu_{\mathcal{B}}$ and $\sigma_{\mathcal{B}}^2$ are the mean and variance of $z$ over a mini-batch. After folding, we obtain new parameters for weights and bias:
% \begin{align}
%     W_{fold} &= \gamma  \frac{W}{\sqrt{\sigma_{\mathcal{B}}^2 + \epsilon}}, \label{eq:folding1}\\
%     b_{fold} &= \gamma  \frac{b-\mu_{\mathcal{B}}}{\sqrt{\sigma_{\mathcal{B}}^2 + \epsilon}} + \beta. \label{eq:folding2}
% \end{align}
% Note that all parameters in Eqs.~(\ref{eq:folding1})\&~(\ref{eq:folding2}) are fixed after training, so the product between flattened weight matrix and learned prototypes can be computed offline. 

For the original {\tt im2col} convolution, the computation complexity is $O(c_{in} H_{out} W_{out}k^2 c_{out})$. During inference, our method includes two stages, the first is to get the indices by computing the distance between the flattened features and prototypes, while the second is to retrieve the product between weights and prototypes computed in advance, i.e., a simple table lookup. The inference algorithm for both PECAN variants is given in Algorithm~\ref{alg:inference}.

Table~\ref{tab:inf_complexity} illustrates the number of multiplication and addition operations in convolution and fully-connected layers for the traditional CNNs, angle-based and distance-based PECAN during the inference phase. Note that the fully-connected layer can be regarded as a convolution layer when $k = H_{out}=W_{out}=1$. Instead of using the specialized setting of $D=c_{in}$ and $d = k^2$, we further consider the more general case in Table~\ref{tab:inf_complexity} where the group number $D$ and dimension of prototypes $d$ satisfy $Dd = c_{in}k^2$. Choosing smaller $p$ and $D$ will reduce the computation complexity for both PECAN-A and PECAN-D. Specifically, in order to limit multiplication complexity in PECAN-A to be smaller than the baseline, we need $p \leq \min (\lambda c_{out}, (1-\lambda)d)$ with $\lambda \in (0,1)$. This constraint is also taken into consideration in the experiment section. Note that by design, PECAN-D needs \emph{no multiplication during inference}, thus making it genuinely totally multiplier-less.

\begin{algorithm}[t]
% \footnotesize
\renewcommand{\algorithmicrequire}{\textbf{Input:}}
\renewcommand{\algorithmicensure}{\textbf{Output:}}
\caption{Inference Algorithm of PECAN}
\label{alg:inference}
\begin{algorithmic}[1]
\REQUIRE Codebook $C \in \mathbb{R}^{c_{in} k^2 \times p}$, 4-D learned kernel tensor $\ten{K} \in \mathbb{R}^{c_{out} \times c_{in} \times k \times k}$, unfolded features $X \in \mathbb{R}^{c_{in} k^2 \times H_{out} W_{out}}$.
\ENSURE The approximated convolution output $\tilde{Y} \in \mathbb{R}^{c_{out} \times H_{out} W_{out}}$
\vspace{1ex}
\STATE Permute and reshape weights to $W_1 \in \mathbb{R}^{D \times c_{out} \times d}$, codebooks to $C_1 \in \mathbb{R}^{D \times d \times p}$
\FOR{ $j$  \textbf{in} $\{1, 2, \cdots, D\}$ }
\STATE $Y^{(j)} = {W_1}^{(j)} {C_1}^{(j)} \in \mathbb{R}^{c_{out} \times p}$
\ENDFOR
\FOR{ $i$  \textbf{in} $\{1, 2, \cdots, H_{out}W_{out}\}$ }
\IF{PECAN-A}
\STATE $\tilde{Y}_i = \sum\limits_{j=1}^D Y^{(j)} {\rm softmax}({C}^{(j)T}X_i^{(j)}) $ %\Comment{PECAN-A}
\ENDIF
\IF{PECAN-D}
\STATE $k_i^{(j)} = \mathop{\arg \max}\limits_{m} -||X_i^{(j)} - {C}_m^{(j)}||_1$
\STATE $\tilde{Y}_i = \sum\limits_{j=1}^D Y^{(j)}_{k_i^{(j)}}$ %\Comment{PECAN-D}
\ENDIF
\ENDFOR
\STATE \textbf{return} Concatenate $(\tilde{Y}_1, \tilde{Y}_2, \cdots, \tilde{Y}_{H_{out} W_{out}})$
\end{algorithmic}
\end{algorithm}

% These two steps have complexity $O(H_{out}  W_{out} p c_{in}$ and $O(c_{out} c_{in} p k^2)$, respectively. Therefore, if we have $p \leq O(\frac{H_{out} W_{out} c_{out} k^2}{H_{out}  W_{out} + c_{out} k^2}) $, we can have keep the computation complexity lower for every layer.

%%%%% Experiments %%%%% 
\section{Experiments}
\label{sec:exp}
To demonstrate the effectiveness of the proposed PECAN and further benchmark the differences between its two variants (PECAN-A and PECAN-D), we apply PECAN to the classification tasks, taking MINST~\cite{deng2012mnist}, CIFAR-10 and CIFAR-100~\cite{krizhevsky2009learning} as datasets. The models employed in this section include modified LeNet5, VGG-Small~\cite{xue2021self}, ResNet20 and ResNet32~\cite{he2016deep}. The experiments explore two PECAN training strategies, namely, co-optimization on weights and codebooks and uni-optimization on codebooks. We also compare PECAN with state-of-the-art (SOTA) approaches that aim to reduce the amount of multiplication operations, namely, XNOR-Net~\cite{rastegari2016xnor}, IR-Net~\cite{qin2020forward}, FDA-BNN~\cite{xu2021learning}, ReCU~\cite{xu2021recu}, SD-BNN~\cite{xue2021self} and AdderNet~\cite{chen2020addernet}. Additionally, we investigate how the dimension and number of groups of the codebook affect the performance of PECAN, validate the necessity of our training strategies, and provide visual results to confirm the approximation capability of the prototypes.

\textbf{Implementation Details.} When using the PECAN framework to train the modified LeNet5 with kernels of size $3\times 3$ on MNIST, we employ the uni-optimization strategy that only updates the prototypes with the trained weights being frozen. The prototypes are trained for $150$ epochs. The learning rate is set to $0.01$ initially, decaying every $50$ epochs. To implement the PECAN framework for the CIFAR-10 and CIFAR-100 tasks, we use the co-optimization strategy and set the training epochs for PECAN-A and PECAN-D as $150$ and $300$, respectively. The learning rate for PECAN-A follows the LeNet5 scheme, while that of PECAN-D is initialized as $0.001$, decaying at epoch $200$. For both datasets, we employ {\tt softmax} function and set the temperature $\tau$ at $1$ and $0.5$ for PECAN-A and PECAN-D, respectively. We set the batch size to $64$, and use cross-entropy as the loss function, which is optimized by Adam. All experiments are run on a machine equipped with four NVIDIA Tesla V100 GPU with 24GB frame buffer, and all codes are implemented by PyTorch.

%We conduct extensive experiments on the image classification tasks, using benchmark datasets including MNIST and CIFAR10, to demonstrate the effectiveness of the proposed PECAN scheme. We mainly compare with the closest schemes AdderNet and BNN which intend to get rid of multiplication operations. The experiments are conducted on NVIDIA Tesla V100 GPU in PyTorch.

\subsection{Modified LeNet5 on MNIST}
To quickly zoom into the superiority of PECAN, the amount of required addition and multiplication operations and the detailed codebook information for each layer in the modified LeNet5 are shown in Appendix Table~\ref{appendix:mnist_exp_setting}. Focusing on the second and third columns, it is noticeable that PECAN-A has fewer multiplications and additions compared with the baseline, and PECAN-D needs no multiplication at all. For the codebook settings, it is seen that the number of prototypes $p$ used in PECAN-A is much fewer than that of PECAN-D for all five layers. We adopt this setting considering the gaps between the representation capabilities of PECAN-A and PECAN-D. By adjusting the weights assigned to prototypes, PECAN-A is expected to better approximate the features with limited choices, i.e., a smaller $p$.

\begin{wraptable}{l}{7cm}
\vspace{-5mm}
\caption{Experiment results of LeNet on MNIST.}
\label{tab:mnist_exp}
\vspace{2mm}
\scriptsize
\centering
\setlength{\tabcolsep}{3mm}{
\begin{tabular}{cccc}
\toprule 
Model & \#Add. & \#Mul. & Acc.(\%) \\
\midrule
Baseline & $248.10K$ & $248.10K$ & $99.41$   \\
PECAN-A & $196.88K$ & $196.88K$ & $99.25$ \\
PECAN-D & $2.00M$ & $0$ & $99.01$ \\
\bottomrule
\end{tabular}}
\vspace{-3mm}
\end{wraptable}

The performance and the required number of addition and multiplication operations of the whole modified LeNet5 employing PECAN-A and PECAN-D schemes are summarized in Table~\ref{tab:mnist_exp}. It is worth noting that LeNet with PECAN-D is multiplier-free, and maintains the good performance compared with the baseline ($99.01\%$ vs $99.41\%$). The accuracy of PECAN-A achieves $99.25\%$, which is merely $0.16\%$ lower than the original LeNet5. However, PECAN-A performs fewer operations. To this end, the LeNet5 example demonstrates the effectiveness of the PECAN framework, and shows the advantages of PECAN-A and PECAN-D from different perspectives.

\begin{table}[h]
\begin{minipage}{.48\linewidth}
\caption{Experiment results on CIFAR10.}
\label{tab:cifar10_exp}
    \scriptsize
    \centering
    \setlength{\tabcolsep}{1mm}{
    \begin{tabular}{ccccc}
    \toprule 
    Model & Method & \#Add. & \#Mul. & Accuracy (\%) \\
    \midrule
    \multirow{3}{*}{VGG-Small} & Baseline & $0.61G$ & $0.61G$ & $91.21$  \\
    ~ & PECAN-A & $0.54G$ & $0.54G$ & $91.82$ \\
    ~ & PECAN-D & $0.37G$ & $0$ & $90.19$ \\
    \midrule
    \multirow{3}{*}{ResNet20} & Baseline & $40.55M$ & $40.55M$ & $92.55$ \\
    ~ & PECAN-A & $38.12M$ & $38.12M$ & $90.32$  \\
    ~ & PECAN-D & $211.71M$ & $0$ & $87.88$ \\
    \midrule
    \multirow{3}{*}{ResNet32} & Baseline & $68.86M$ & $68.86M$ & $92.85$  \\
    ~ & PECAN-A & $64.20M$ & $64.20M$ & $90.53$  \\
    ~ & PECAN-D & $353.26M$ & $0$ & $88.46$ \\
    \bottomrule
\end{tabular}}
\end{minipage}
% \\[12pt]
\begin{minipage}{0.48\linewidth}
\centering
\scriptsize
\caption{Experiment results on CIFAR100.}
\label{tab:cifar100_exp}
\scriptsize
\centering
\setlength{\tabcolsep}{1mm}{
\begin{tabular}{ccccc}
\toprule 
Model & Method & \#Add. & \#Mul. & Accuracy (\%) \\
\midrule
\multirow{3}{*}{VGG-Small} & Baseline & $0.61G$ & $0.61G$ & $67.84$  \\
~ & PECAN-A & $0.54G$ & $0.54G$ & $69.21$ \\
~ & PECAN-D & $0.37G$ & $0$ & $60.43$ \\
\midrule
\multirow{3}{*}{ResNet20} & Baseline & $40.56M$ & $40.56M$ &  $69.55$ \\
~ & PECAN-A & $38.12M$ & $38.12M$ & $63.15$  \\
~ & PECAN-D & $211.71M$ & $0$ &  $58.01$ \\
\midrule
\multirow{3}{*}{ResNet32} & Baseline & $68.86M$ & $68.86M$ & $70.57$  \\
~ & PECAN-A & $64.20M$ & $64.20M$ &  $64.13$ \\
~ & PECAN-D & $353.27M$ & $0$ &  $58.26$ \\
\bottomrule
\end{tabular}}
\end{minipage}
\vspace{-3mm}
\end{table}

%------------comparison with BNN ----------------
% \caption{Comparison with SOTA BNNs.}
% \label{tab:comparison_cifar10_bnn}
% % \vspace{1mm}
%     \centering
%     \scriptsize
%      \setlength{\tabcolsep}{0.8mm}{
%     \begin{tabular}{cccc}
%     \toprule
%         Model & Method & Bit-width (W/A) & Accuracy(\%) \\
%     \midrule
%         \multirow{5}{*}{ResNet-20} & Baseline & 32/32  & $92.55$ \\
%         ~ & XNOR-Net~\cite{rastegari2016xnor} & 1/1 & $85.23$ \\
%         ~ & IR-Net~\cite{qin2020forward} & 1/1 & $85.40$ \\
%         ~ & FDA-BNN~\cite{xu2021learning} & 1/1 & $86.20$ \\
%         ~ & ReCU~\cite{xu2021recu} & 1/1 & $87.4$ \\
%         ~ & PECAN-D & 32/32 & $87.88$ \\
%     \midrule
%         \multirow{5}{*}{VGG-Small} & Baseline & 32/32  & $91.21$ \\
%         ~ & XNOR-Net & 1/1 & $89.80$ \\
%         ~ & IR-Net & 1/1 & $90.40$ \\
%         ~ & SD-BNN~\cite{xue2021self} & 1/1 & $90.8$ \\
%         ~ & PECAN-D & 32/32 & $90.19$ \\
%     \bottomrule
%     \end{tabular}}

\subsection{VGG and ResNet on CIFAR-10/100}
After the proof-of-concept on LeNet5, we proceed to VGG-Small and ResNet20/32 on CIFAR-10 and CIFAR-100. VGG-Small is a simplified VGGNet~\cite{simonyan2014very} with only one fully-connected layer. The size of the output feature maps and the corresponding codebook information for each layer are provided in Appendix Table~\ref{appendix:p_setting_cifar10}. We remark that the bottom row of each block in the table represents the FC layer, while the rows above represent the CONV layers. The number of required addition and multiplication operations and the accuracy of the models are summarized in Table~\ref{tab:cifar10_exp}, where it can seen that the VGG-Small baseline has $0.61G$ multiplication and addition operations with $91.21\%$ accuracy. Since batch normalization can be folded into convolution layers in the inference stage, we do not count FLOPs for both baseline and PECAN. We find that PECAN-A only performs $0.54G$ multiplications while reaching $91.82\%$ accuracy on CIFAR-10, which is even higher than the baseline, similar performance can be obtained on CIFAR-100 in Table~\ref{tab:cifar100_exp}. A possible reason is that PECAN experiences less information loss for shallower CNNs, and bigger input channels allow more groups of prototypes to improve the representation capability. This assumption is also validated by the experiments on ResNet20/32 that are deeper than VGG-Small but with smaller input channels.

Although PECAN-D has an accuracy drop compared with the baseline, it eliminates all multiplications during inference. For VGG-Small, the number of additions reduces to $0.37G$ while accuracy drops only around $1\%$ compared with the baseline on CIFAR10. To boost the performance, we use smaller size for subvectors in PECAN-D as shown in the last column of Appendix Table~\ref{appendix:p_setting_cifar10}, at the expense of more computation. 

%%%%%%%%%%% subsection: SOTA --> only AdderNet
\subsection{Comparison with AdderNet}
We compare PECAN-D with AdderNet on VGG-Small in Table~\ref{tab:comparison_cifar10_adder}. It should be emphasized that batch normalization is not taken into consideration in this table, it can not be folded into AdderNet layer so multiplication is indispensable. For VGG-Small, the memory cost is so high that even four NVIDIA Tesla V100 GPUs are not able to train successfully. As shown in the table, the proposed PECAN-D with only ~0.37G additions achieves a $90.19\%$ accuracy on VGG-Small. Here we showcase the efficacy of PECAN on larger models from a hardware perspective. In the Intel VIA Nano 2000 CPU (used in the AdderNet paper), the latency cycles of float multiplication and addition are 4 and 2, respectively. PECAN-D of VGG-Small model will incur $\sim$720M(cycles) while that of a CNN is $\sim$3660M. The power consumption ratio of 32bit multiplication and addition units is 4:1. Power-wise and latency-wise, PECAN-D network is more efficient than both AdderNet and regular CNN.

% -------table of comparison with AdderNet -------
\begin{table}[h]
\caption{Comparison with AdderNet.}
    \centering
    \scriptsize
    \setlength{\tabcolsep}{2mm}{
    \begin{tabular}{ccccccc}
    \toprule
        Model & Method & \# Mul. & \# Add. & Accuracy (\%) & Normalized Power & Latency(cycles)\\
    \midrule
        \multirow{3}{*}{VGG-Small} & CNN & $0.61G$ & $0.61G$ & $93.80$ & $8.24$ & $3.66G$\\
        ~ & AdderNet & $0$ & $1.22G$ & N.A. & $3.30$ & $2.44G$\\
        ~ & PECAN-D & $0$ & $0.37G$ & $90.19$ & $1$ & $0.72G$\\
    % \midrule
    %     \multirow{3}{*}{ResNet20} & CNN & $40.55M$ & $40.55M$ & $92.55$ & $0.202$ & $0.243G$ \\
    %     ~ & AdderNet & $0.44M$ & $80.66M$ & $90.87$ & $0.0824$ & $0.163G$ \\
    %     ~ & PECAN-D & $0$ & $211.71M$ & $87.88$ & $0.211$ & $0.423G$ \\
    \bottomrule
    \end{tabular}
    \label{tab:comparison_cifar10_adder}}
\end{table}
%Based on the image classification tasks on CIFAR10, we further evaluate PECAN-D with VGG-Small and ResNet20 by comparing it with existing SOTA methods that attempt to get rid of most multiplications, including XNOR-Net~\cite{rastegari2016xnor}, IR-Net~\cite{qin2020forward}, FDA-BNN~\cite{xu2021learning}, ReCU~\cite{xu2021recu}, SD-BNN~\cite{xue2021self} and AdderNet~\cite{chen2020addernet}.

% \end{table}
% Table~\ref{tab:comparison_cifar10_bnn} shows the comparison results with SOTA BNNs. PECAN-D outperforms other BNN methods on ResNet20 with accuracy $87.88\%$. It is worth noting all methods other than PECAN-D retain full-precision first (convolution) and last (fully-connected) layers, and only PECAN-D is truly multiplication-free. Therefore, it is reasonable that a narrow accuracy gap exists for VGG-Small versus IR-Net and SD-BNN.

%%%% ablation study
\subsection{Ablation Study}
% We conduct detailed ablation studies on prototype dimensions and keeping the weights frozen during fine-tuning for PECAN-D. For brevity, we report ResNet20 and VGG-Small on CIFAR10 in Table~\ref{tab:cifar10_exp}, while we can observe similar results from other networks and datasets.

%%%% fig: ablation last

\subsubsection{Prototypes Dimension}
%%% fig: ablation first
\begin{wrapfigure}{r}{7cm}
    \vspace{-2mm}
    \centering
    \includegraphics[scale=0.4]{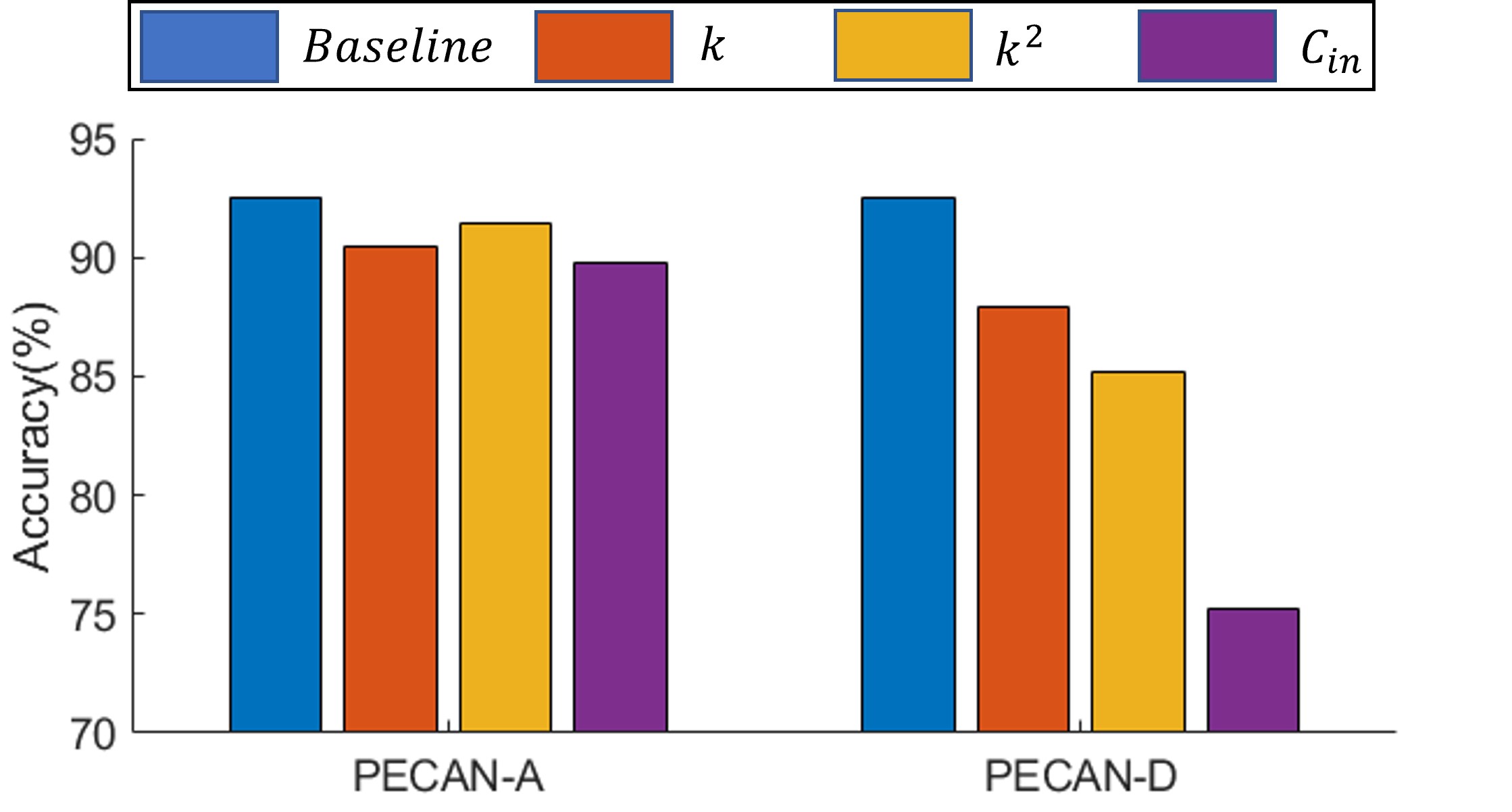}
    \vspace{-2mm}
    \caption{Accuracy of ResNet20 on CIFAR10 using different dimensions of subvectors including $k$, $k^2$ and $C_{in}$ for both PECAN-A and PECAN-D.}
    \label{fig:dim_pro}
\end{wrapfigure}
To investigate the effect of the prototype dimension on the performance of both PECAN-A and PECAN-D, we conduct comparative experiments on ResNet20 (on CIFAR-10 dataset) by reducing the dimension from $k^2$ to $k$, and then increasing up to $c_{in}$. Subsequently, the number of groups changes from $c_{in}$ to $k c_{in}$ and $k^2$ correspondingly. Except for the prototype dimension, we keep other settings (e.g., number of prototypes $p$ in each layer, and learning rate decay scheme) the same as the experiments in Table~\ref{tab:cifar10_exp}. The results are visualized in Fig.~\ref{fig:dim_pro}, wherein the prototype dimension increases from left to right for each set of bar charts. It is observed that the performances of PECAN-A and PECAN-D have different trends when the prototype dimension increases. Compared with PECAN-D, the angle-based PECAN-A is more robust, and its accuracy does not change sharply under the three cases. It achieves the best performance when the prototype dimension is $k^2$, while still maintaining decent accuracy when the dimension changes to $k$ and $c_{in}$. On the contrary, the performance of PECAN-D is more sensitive, which is inversely proportional to the prototype dimension. It is intuitive since approximation in the fine-scale (viz. larger groups with smaller dimensions) is expected to be more accurate. We remark that PECAN-A, which has weighted prototypes, trades for robustness at all scales by the higher computational complexity compared with PECAN-D.
%As Fig.~\ref{fig:gemm} shows, we divide the flattened features into $c_{in}$ groups so each prototype is of dimension $k^2$. As the kernel size is mostly set as $3, 5, 7$, etc., the dimension of prototypes is relatively small compared with the number of input channels. Therefore, we swap $c_{in}$ and $k^2$ which means the prototypes are divided into $k^2$ groups only and the dimension of them extends to $c_{in}$. Smaller dimension $k$ is also added into comparison with $k C_{in}$ groups. We report experimental results in Fig.~\ref{fig:dim_pro} and keep other settings including the number of prototypes in each layer and the training parameters the same as Table~\ref{tab:cifar10_exp}.

%From the results, we can find when the dimension of prototypes increases, the performance of PECAN-D drops accordingly. This is reasonable because the templates match better using a single prototype when divided into more groups. On the contrary, PECAN-A is more robust with respect to the length of prototypes, those with dimension $k^2$ achieves higher accuracy than the other scenarios. 

\subsubsection{Freezing Weights during Training}

For both angle- and distance-based measures, we freeze the pretrained weights and only train the matched codebooks in the MNIST example, but when moving to a larger dataset, both weights and prototypes are not frozen but are updated during training from scratch. To further clarify the reason of this choice, we also train VGG-Small which has the same setting as in Table~\ref{tab:cifar10_exp} on CIFAR10. 

\begin{wraptable}{l}{7cm}
\centering
\scriptsize
\vspace{-5mm}
\caption{Effects of training strategies on PECAN accuracy.}
\vspace{2mm}
 \setlength{\tabcolsep}{1mm}{
\begin{tabular}{cccc}
\toprule
    Model & From Scratch & Freeze Weights & Accuracy(\%) \\
\midrule
    Baseline & \Checkmark & \XSolidBrush & $91.21$ \\
    PECAN-A/D & \Checkmark & \XSolidBrush & $91.82/90.19$ \\
    PECAN-A/D & \XSolidBrush & \Checkmark & $91.76/87.43$ \\
\bottomrule
\end{tabular}}
\vspace{-2mm}
\label{tab:freeze_or_not}
\end{wraptable}
However, different from training from scratch, we only unfreeze prototypes and start from the pretrained mature CNN. Specifically, we initialize all convolution filters with the pretrained model and only learn those prototypes in each layer. Experimental results are shown in Table~\ref{tab:freeze_or_not}. As seen from the table, updating prototypes only still has accuracy gap especially for PECAN-D compared with the one learning both weights and prototypes from scratch. A possible reason is that the convolution weights in the pretrained model are not well-matching with the templates.

\subsubsection{Visualization of Prototypes}
\begin{figure*}[t]
\centering
\includegraphics[scale=0.4]{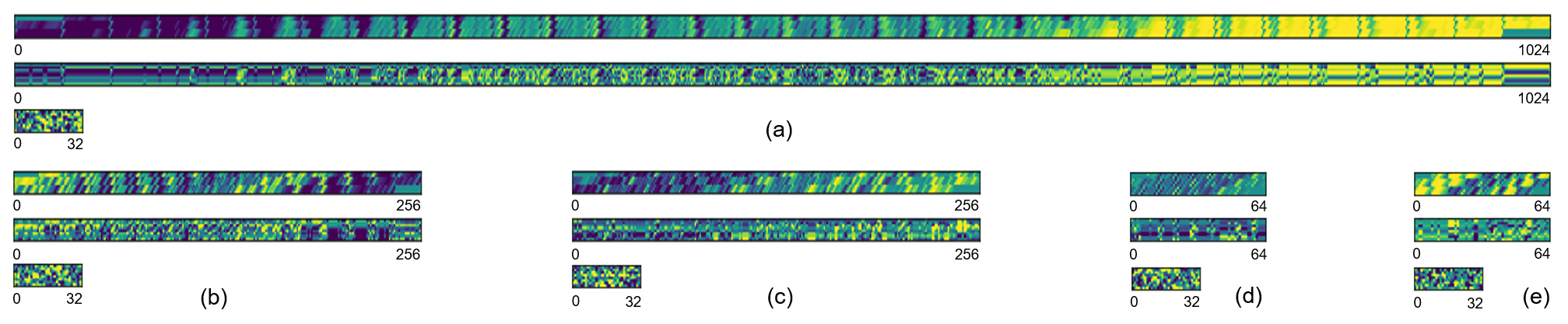}
\vspace{-1mm}
\caption{The flattened features and codebooks for five different layers in VGG-Small, (a)-(e) for conv1-conv5. For each subfigure, the upper image is the input feature after {\tt{im2col}} operation, the second image shows the approximation matrix after substitution with PECAN-D which is composed of the corresponding codebook shown in the third row. The $y$-axis is the dimension of each subvector $k^2$. The $x$-axis represents the size of output feature maps $H_{out}W_{out}$ for the first two rows, and denotes the number of prototypes for the third row.}
\label{fig:vggsmall_in_out}
\end{figure*}

To visually inspect the effectiveness of PECAN-D in CNNs, we take the intermediate convolution layers of VGG-Small and plot the patterns of the feature maps before and after replacement. In Fig.~\ref{fig:vggsmall_in_out}, we select the first channel of the flattened feature maps and visualize the matrices. The dimension of all subvectors is set as $k^2 = 9$. As can be seen, though the number of prototypes is limited for each convolution layer, the quantized feature maps can still preserve the basic patterns after training.

%%%%%%%% Discussion %%%%%%%%%
\section{Discussion}

\begin{wrapfigure}{r}{7cm}
    \centering
    \vspace{-4mm}
    \includegraphics[scale=0.58]{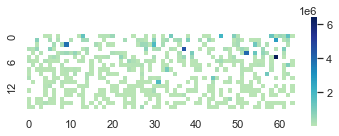}
    \vspace{-4mm}
    \caption{Call frequencies of 64 prototypes in the middle 18 CNN layers @ ResNet20. The $x$-axis represents the indices of prototypes and the $y$-axis is the order of the 18 layers. White grid cells denote 0 times of usage.}
    \vspace{-2mm}
    \label{fig:prototype_idx}
\end{wrapfigure}
To boost the performance for PECAN, we can choose small size of prototypes in our experiments which might incur a challenge for memory footprint. Reducing the number of prototypes and increasing the subvector lengths can lower the memory cost, but this may harm the model accuracy. To this end, We propose further means to save memory: For PECAN-A we can exploit the projectors inherent to attention to shrink dimensions of query and key (corresponding to weights \& prototypes in PECAN-A). Even more interesting, for PECAN-D, take the 2nd CONV layer of ResNet20 on CIFAR10 as an example, only 26 out of 64 prototypes are used in the inference stage, meaning all other prototypes and lookup entries can be pruned without affecting accuracy. Fig.~\ref{fig:prototype_idx} illustrates the sparse usage count of each prototype in the first group of codebooks for 18 intermediate CNN layers. We will report these exciting results in the follow-up work of PECAN, as these are beyond the central theme of this paper.

%%%%% Conclusion %%%%% 
\section{Conclusion}
A brand new DNN architecture called PECAN is proposed which transcends the regular DNN linear transform, and replaces it by product quantization and table lookup. Both angle- and distance-based measures are developed for similarity matching of prototypes in product quantization for different complexity-accuracy tradeoffs. The distance-based PECAN, to our knowledge, is the \emph{first} neural network that is multiplier-less and uses only adders all over. PECAN is end-to-end trainable and infers only through a content addressable memory (CAM)-like, similarity search protocol. It facilitates a lightweight and hardware-generic solution favorable for edge AI, and fits perfectly into the in-memory-computing regime. Experiments have shown that PECAN exhibits accuracies on par with multi-bit networks even without using multipliers. We expect more advancement on top of this interesting PECAN framework will follow after this debut.%
\label{sec:conclusion}

%%%%%%%%% REFERENCES
\clearpage
\bibliographystyle{plainnat}
\bibliography{egbib}

\begin{thebibliography}{30}
\providecommand{\natexlab}[1]{#1}
\providecommand{\url}[1]{\texttt{#1}}
\expandafter\ifx\csname urlstyle\endcsname\relax
  \providecommand{\doi}[1]{doi: #1}\else
  \providecommand{\doi}{doi: \begingroup \urlstyle{rm}\Url}\fi

\bibitem[Blalock and Guttag(2021)]{blalock2021multiplying}
Davis Blalock and John Guttag.
\newblock Multiplying matrices without multiplying.
\newblock \emph{arXiv preprint arXiv:2106.10860}, 2021.

\bibitem[Chen et~al.(2020{\natexlab{a}})Chen, Wang, Xu, Shi, Xu, Tian, and
  Xu]{chen2020addernet}
Hanting Chen, Yunhe Wang, Chunjing Xu, Boxin Shi, Chao Xu, Qi~Tian, and Chang
  Xu.
\newblock Addernet: Do we really need multiplications in deep learning?
\newblock In \emph{Proceedings of the IEEE/CVF Conference on Computer Vision
  and Pattern Recognition}, pages 1468--1477, 2020{\natexlab{a}}.

\bibitem[Chen et~al.(2020{\natexlab{b}})Chen, Li, and
  Sun]{chen2020differentiable}
Ting Chen, Lala Li, and Yizhou Sun.
\newblock Differentiable product quantization for end-to-end embedding
  compression.
\newblock In \emph{International Conference on Machine Learning}, pages
  1617--1626. PMLR, 2020{\natexlab{b}}.

\bibitem[Courbariaux et~al.(2015)Courbariaux, Bengio, and
  David]{courbariaux2015binaryconnect}
Matthieu Courbariaux, Yoshua Bengio, and Jean-Pierre David.
\newblock Binaryconnect: Training deep neural networks with binary weights
  during propagations.
\newblock In \emph{Advances in neural information processing systems}, pages
  3123--3131, 2015.

\bibitem[Deng(2012)]{deng2012mnist}
Li~Deng.
\newblock The mnist database of handwritten digit images for machine learning
  research.
\newblock \emph{IEEE Signal Processing Magazine}, 29\penalty0 (6):\penalty0
  141--142, 2012.

\bibitem[Dong et~al.(2021)Dong, Wang, Chen, and Xu]{dong2021towards}
Minjing Dong, Yunhe Wang, Xinghao Chen, and Chang Xu.
\newblock Towards stable and robust addernets.
\newblock In \emph{Thirty-Fifth Conference on Neural Information Processing
  Systems}, 2021.

\bibitem[Elhoushi et~al.(2021)Elhoushi, Chen, Shafiq, Tian, and
  Li]{elhoushi2021deepshift}
Mostafa Elhoushi, Zihao Chen, Farhan Shafiq, Ye~Henry Tian, and Joey~Yiwei Li.
\newblock Deepshift: Towards multiplication-less neural networks.
\newblock In \emph{Proceedings of the IEEE/CVF Conference on Computer Vision
  and Pattern Recognition}, pages 2359--2368, 2021.

\bibitem[Gudovskiy and Rigazio(2017)]{gudovskiy2017shiftcnn}
Denis~A Gudovskiy and Luca Rigazio.
\newblock Shiftcnn: Generalized low-precision architecture for inference of
  convolutional neural networks.
\newblock \emph{arXiv preprint arXiv:1706.02393}, 2017.

\bibitem[He et~al.(2016)He, Zhang, Ren, and Sun]{he2016deep}
Kaiming He, Xiangyu Zhang, Shaoqing Ren, and Jian Sun.
\newblock Deep residual learning for image recognition.
\newblock In \emph{Proceedings of the IEEE conference on computer vision and
  pattern recognition}, pages 770--778, 2016.

\bibitem[Hubara et~al.(2016)Hubara, Courbariaux, Soudry, El-Yaniv, and
  Bengio]{hubara2016binarized}
Itay Hubara, Matthieu Courbariaux, Daniel Soudry, Ran El-Yaniv, and Yoshua
  Bengio.
\newblock Binarized neural networks.
\newblock \emph{Advances in neural information processing systems}, 29, 2016.

\bibitem[Jegou et~al.(2010)Jegou, Douze, and Schmid]{jegou2010product}
Herve Jegou, Matthijs Douze, and Cordelia Schmid.
\newblock Product quantization for nearest neighbor search.
\newblock \emph{IEEE transactions on pattern analysis and machine
  intelligence}, 33\penalty0 (1):\penalty0 117--128, 2010.

\bibitem[Karunaratne et~al.(2021)Karunaratne, Schmuck, Le~Gallo, Cherubini,
  Benini, Sebastian, and Rahimi]{Karunaratne2021RobustHM}
Geethan Karunaratne, Manuel Schmuck, Manuel Le~Gallo, Giovanni Cherubini, Luca
  Benini, Abu Sebastian, and Abbas Rahimi.
\newblock Robust high-dimensional memory-augmented neural networks.
\newblock \emph{Nat. Commun.}, 12, 2021.

\bibitem[Krizhevsky et~al.(2009)Krizhevsky, Hinton,
  et~al.]{krizhevsky2009learning}
Alex Krizhevsky, Geoffrey Hinton, et~al.
\newblock Learning multiple layers of features from tiny images.
\newblock 2009.

\bibitem[Le and Yang(2015)]{le2015tiny}
Ya~Le and Xuan Yang.
\newblock Tiny imagenet visual recognition challenge.
\newblock \emph{CS 231N}, 7\penalty0 (7):\penalty0 3, 2015.

\bibitem[Liu et~al.(2020)Liu, Wang, Wang, Liang, Zhao, Tang, and
  Ling]{liu2020cbnet}
Yudong Liu, Yongtao Wang, Siwei Wang, TingTing Liang, Qijie Zhao, Zhi Tang, and
  Haibin Ling.
\newblock Cbnet: A novel composite backbone network architecture for object
  detection.
\newblock In \emph{Proceedings of the AAAI conference on artificial
  intelligence}, volume~34, pages 11653--11660, 2020.

\bibitem[Qin et~al.(2020)Qin, Gong, Liu, Shen, Wei, Yu, and
  Song]{qin2020forward}
Haotong Qin, Ruihao Gong, Xianglong Liu, Mingzhu Shen, Ziran Wei, Fengwei Yu,
  and Jingkuan Song.
\newblock Forward and backward information retention for accurate binary neural
  networks.
\newblock In \emph{Proceedings of the IEEE/CVF Conference on Computer Vision
  and Pattern Recognition}, pages 2250--2259, 2020.

\bibitem[Rahimi et~al.(2007)Rahimi, Recht, et~al.]{rahimi2007random}
Ali Rahimi, Benjamin Recht, et~al.
\newblock Random features for large-scale kernel machines.
\newblock In \emph{NIPS}, volume~3, page~5. Citeseer, 2007.

\bibitem[Rastegari et~al.(2016)Rastegari, Ordonez, Redmon, and
  Farhadi]{rastegari2016xnor}
Mohammad Rastegari, Vicente Ordonez, Joseph Redmon, and Ali Farhadi.
\newblock Xnor-net: Imagenet classification using binary convolutional neural
  networks.
\newblock In \emph{European conference on computer vision}, pages 525--542.
  Springer, 2016.

\bibitem[Ren et~al.(2021)Ren, Lin, Ran, Liu, Tao, Wang, Li, and Wong]{batmann}
Yuan Ren, Rui Lin, Jie Ran, Chang Liu, Chaofan Tao, Zhongrui Wang, Can Li, and
  Ngai Wong.
\newblock Batmann: A binarized-all-through memory-augmented neural network for
  efficient in-memory computing.
\newblock In \emph{2021 IEEE 14th International Conference on ASIC (ASICON)},
  pages 1--4, 2021.
\newblock \doi{10.1109/ASICON52560.2021.9620292}.

\bibitem[Simonyan and Zisserman(2014)]{simonyan2014very}
Karen Simonyan and Andrew Zisserman.
\newblock Very deep convolutional networks for large-scale image recognition.
\newblock \emph{arXiv preprint arXiv:1409.1556}, 2014.

\bibitem[Sun et~al.(2020)Sun, Xue, Zhang, Yen, and Lv]{sun2020automatically}
Yanan Sun, Bing Xue, Mengjie Zhang, Gary~G Yen, and Jiancheng Lv.
\newblock Automatically designing cnn architectures using the genetic algorithm
  for image classification.
\newblock \emph{IEEE transactions on cybernetics}, 50\penalty0 (9):\penalty0
  3840--3854, 2020.

\bibitem[Trockman and Kolter(2022)]{trockman2022patches}
Asher Trockman and J~Zico Kolter.
\newblock Patches are all you need?
\newblock \emph{arXiv preprint arXiv:2201.09792}, 2022.

\bibitem[Vaswani et~al.(2017)Vaswani, Shazeer, Parmar, Uszkoreit, Jones, Gomez,
  Kaiser, and Polosukhin]{vaswani2017attention}
Ashish Vaswani, Noam Shazeer, Niki Parmar, Jakob Uszkoreit, Llion Jones,
  Aidan~N Gomez, {\L}ukasz Kaiser, and Illia Polosukhin.
\newblock Attention is all you need.
\newblock In \emph{Advances in neural information processing systems}, pages
  5998--6008, 2017.

\bibitem[Xu et~al.(2020)Xu, Xu, Chen, Zhang, Xu, and Wang]{xu2020kernel}
Yixing Xu, Chang Xu, Xinghao Chen, Wei Zhang, Chunjing Xu, and Yunhe Wang.
\newblock Kernel based progressive distillation for adder neural networks.
\newblock \emph{arXiv preprint arXiv:2009.13044}, 2020.

\bibitem[Xu et~al.(2021{\natexlab{a}})Xu, Han, Xu, Tang, Xu, and
  Wang]{xu2021learning}
Yixing Xu, Kai Han, Chang Xu, Yehui Tang, Chunjing Xu, and Yunhe Wang.
\newblock Learning frequency domain approximation for binary neural networks.
\newblock \emph{arXiv preprint arXiv:2103.00841}, 2021{\natexlab{a}}.

\bibitem[Xu et~al.(2021{\natexlab{b}})Xu, Lin, Liu, Chen, Shao, Gao, Tian, and
  Ji]{xu2021recu}
Zihan Xu, Mingbao Lin, Jianzhuang Liu, Jie Chen, Ling Shao, Yue Gao, Yonghong
  Tian, and Rongrong Ji.
\newblock Recu: Reviving the dead weights in binary neural networks.
\newblock \emph{arXiv preprint arXiv:2103.12369}, 2021{\natexlab{b}}.

\bibitem[Xue et~al.(2021)Xue, Lu, Chang, Wei, and Wei]{xue2021self}
Ping Xue, Yang Lu, Jingfei Chang, Xing Wei, and Zhen Wei.
\newblock Self-distribution binary neural networks.
\newblock \emph{arXiv preprint arXiv:2103.02394}, 2021.

\bibitem[You et~al.(2020)You, Chen, Zhang, Li, Li, Liu, Wang, and
  Lin]{you2020shiftaddnet}
Haoran You, Xiaohan Chen, Yongan Zhang, Chaojian Li, Sicheng Li, Zihao Liu,
  Zhangyang Wang, and Yingyan Lin.
\newblock Shiftaddnet: A hardware-inspired deep network.
\newblock \emph{arXiv preprint arXiv:2010.12785}, 2020.

\bibitem[Zheng et~al.(2021)Zheng, Lu, Zhao, Zhu, Luo, Wang, Fu, Feng, Xiang,
  Torr, et~al.]{zheng2021rethinking}
Sixiao Zheng, Jiachen Lu, Hengshuang Zhao, Xiatian Zhu, Zekun Luo, Yabiao Wang,
  Yanwei Fu, Jianfeng Feng, Tao Xiang, Philip~HS Torr, et~al.
\newblock Rethinking semantic segmentation from a sequence-to-sequence
  perspective with transformers.
\newblock In \emph{Proceedings of the IEEE/CVF Conference on Computer Vision
  and Pattern Recognition}, pages 6881--6890, 2021.

\bibitem[Zhou et~al.(2017)Zhou, Yao, Guo, Xu, and Chen]{zhou2017incremental}
Aojun Zhou, Anbang Yao, Yiwen Guo, Lin Xu, and Yurong Chen.
\newblock Incremental network quantization: Towards lossless cnns with
  low-precision weights.
\newblock \emph{arXiv preprint arXiv:1702.03044}, 2017.

\end{thebibliography}

\clearpage
\appendix
%------------- Appendix -------------
\appendix
\setcounter{page}{1}
\renewcommand\thefigure{A\arabic{figure}}    
\setcounter{figure}{0}    
\renewcommand\thetable{A\arabic{table}}
\setcounter{table}{0}
\setcounter{equation}{0}
\renewcommand\theequation{A\arabic{equation}}

%----------Details of computation complexity -----------
\section{Computation Complexity of PECAN}
In this section, we analyze the complexity of PECAN-A and PECAN-D during the inference stage, which is shown in Table~\ref{tab:inf_complexity}. For both PECAN-A and PECAN-D, there are two stages of calculation: 1) computing the distance between the flattened features and prototypes,  and 2) a simple table lookup to retrieve the product between weights and prototypes computed in advance.

Specifically, the first stage requires $H_{out}W_{out}$ subvectors in each group to compare with $p$ prototypes. Therefore, PECAN-A needs $H_{out}W_{out}Dp\cdot d$ multiplications and additions, respectively, while PECAN-D has $H_{out}W_{out}Dp\cdot 2d$ additions. During the second stage, a lookup table of $c_{out} \times Dp$ is available to address the quantized product. It takes the weighted sum for PECAN-A or summation for PECAN-D in $D$ groups. We can get $H_{out}W_{out}Dpc_{out}$ additions and multiplications for PECAN-A and $H_{out}W_{out}Dc_{out}$ for PECAN-D.

Since the FC layer can be regarded as a CONV layer when 
$k = H_{out}=W_{out}=1$, the computation complexity of an FC layer can be obtained accordingly.

%-------------Details of PECAN ------------------
\section{Details of PECAN on MNIST}

\setlength{\tabcolsep}{1.4mm}{
\begin{table}[h]
\caption{Detail structure of LeNet used in PECAN.}
\label{appendix:PECAN_lenet}
\vspace{2mm}
\scriptsize
\centering
\begin{tabular}{llclcl}
\toprule
\multirow{2}{*}{LeNet} & kernel size & ~ & Output & ~ & Flattened Weights  \\
\cline{2-2}
\cline{4-4}
\cline{6-6}
~ & $k \times k$ & ~ & $[c_{out}, H_{out}, W_{out}]$ & ~ & $[c_{out}, k^2 c_{in}]$\\
\midrule
CONV1 & $3 \times 3$ & ~ & $[8, 26, 26]$ & ~ & $[8, 9]$ \\
ReLU1 & - & ~ & $[8, 26, 26]$ & ~ & $-$\\
MaxPooling1 & $2 \times 2$ & ~ & $[8, 13, 13]$ & ~ & $-$\\
\midrule
CONV2 & $3 \times 3$ & ~ & $[16, 11, 11]$ & ~ & $[16, 72]$\\
ReLU2 & - & ~ & $[16, 11, 11]$ & ~ & $-$\\
MaxPooling2 & $2 \times 2$ & ~ & $[16, 5, 5]$ & ~ & $-$\\
\midrule
FC1 & $1 \times 1$ & ~ & $[128, 1, 1]$ & ~ & $[128, 400]$ \\
ReLU3 & - & ~ & $[128, 1, 1]$ & ~ & $-$\\
\midrule
FC2 & $1 \times 1$ & ~ & $[64, 1, 1]$ & ~ & $[64, 128]$\\
ReLU4 & - & ~ & $[64, 1, 1]$ & ~ & $-$\\
\midrule
FC3 & $1 \times 1$ & ~ & $[10, 1, 1]$ & ~ & $[10, 64]$\\
\bottomrule
\end{tabular}
\end{table}}

% tab: mnist setting
\setlength{\tabcolsep}{2.7mm}{
\begin{table}[h]
\caption{PECAN settings of LeNet on MNIST.}
\label{appendix:mnist_exp_setting}
\vspace{1mm}
\scriptsize
\centering
\begin{tabular}{lccccc}
\toprule 
Layer & \#Add. & \#Mul. & $p$ & $D$ & $d$\\
\midrule
CONV1 & $48.67K$ & $48.67K$ & -  & - & -\\
CONV1(PECAN-A) & $45.97K$  & $45.97K$ & $4$ & $1$ & $9$ \\
CONV1(PECAN-D) & $784.16K$ & $0$ & $64$ & $1$ & $9$\\
\midrule
CONV2 & $139.39K$ & $139.39K$ & - & - & - \\
CONV2(PECAN-A) & $116.16K$ & $116.16K$ & $8$ & $3$ & $24$\\
CONV2(PECAN-D) & $1.13M$ & $0$ & $64$ & $8$ & $9$\\
\midrule
FC1 & $51.2K$ & $51.2K$ & - & - & - \\
FC1(PECAN-A) & $28.8K$ & $28.8K$ & $8$ & $25$ & $16$ \\
FC1(PECAN-D) & $57.60K$ & $0$ & $64$ & $50$ & $8$ \\
\midrule
FC2 & $8.19K$ & $8.19K$ & - & - & - \\
FC2(PECAN-A) & $5.12K$ & $5.12K$ & $8$ & $8$ & $16$ \\
FC2(PECAN-D) & $17.41K$ & $0$ & $64$ & $16$ & $8$ \\
\midrule
FC3 & $0.64K$ & $0.64K$ & - & - & - \\
FC3(PECAN-A) & $0.83K$ & $0.83K$ & $8$ & $4$ & $16$ \\
FC3(PECAN-D) & $8.27K$ & $0$ & $64$ & $8$ & $8$ \\
\bottomrule
\end{tabular}
\end{table}}

\section{Details of PECAN on CIFAR10}
For PECAN, specialized settings are employed for different models. Table~\ref{appendix:p_setting_cifar10} describes detailed information for each layer in VGG-Small and ResNet20/32.

%%% tab: CIFAR10 setting
\begin{table}[h]
\centering
\scriptsize
\caption{The settings of prototype numbers and dimensions for each layer in different models for PECAN on CIFAR10.}
\label{appendix:p_setting_cifar10}
\vspace{1mm}
\setlength{\tabcolsep}{1mm}{
\begin{tabular}{ccccc}
\toprule
Model & \#Layers & Output map size & $p/d$ (PECAN-A) & $p/d$ (PECAN-D) \\
\midrule
\multirow{4}{*}{VGG-Small} & $2$ & $32 \times 32$ & $16/9$ & $32/3$\\
~ & $2$ & $16 \times 16$ & $16/32$ & $32/3$ \\
~ & $2$ & $8 \times 8$ & $16/32$ & $32/3$ \\
~ & $1$ & $1 \times 1$ & $16/16$ & $32/16$ \\
\midrule
\multirow{4}{*}{ResNet20} & $1$ & $32 \times 32$ & $8/9$ & $128/3$\\
~ & $6$ & $32 \times 32$ & $8/9$ & $64/3$\\
~ & $6$ & $16 \times 16$ & $8/16$ & $64/3$ \\
~ & $6$ & $8 \times 8$ & $8/16$ & $64/3$ \\
~ & $1$ & $1 \times 1$ & $8/16$ & $64/4$ \\
\midrule
\multirow{4}{*}{ResNet32} & $1$ & $32 \times 32$ & $8/9$ & $128/3$\\
~ & $10$ & $32 \times 32$ & $8/9$ & $64/3$\\
~ & $10$ & $16 \times 16$ & $8/16$ & $64/3$ \\
~ & $10$ & $8 \times 8$ & $8/16$ & $64/3$ \\
~ & $1$ & $1 \times 1$ & $8/16$ & $64/4$ \\
\bottomrule
\end{tabular}}
\end{table}

\section{Additional Experiments on Tiny-ImageNet}
Due to space limitations, we put the experimental results on a larger dataset TinyImageNet~\cite{le2015tiny} here. Different from the main paper, we choose ConvMixer~\cite{trockman2022patches} replacing all pointwise and depthwise convolution layers with conventional convolution layers. Besides, we keep the first convolution layer and the last fully-connected layer uncompressed. The depth of ConvMixer is $8$ and kernel sizes in all blocks are $k = 5$. We set $p=16$, $d=25$ for PECAN-A and $p=32$, $d=25$ for PECAN-D. As shown in Table~\ref{tab:tinyimagenet_exp}, PECAN-A achieves $59.42\%$ accuracy which is much higher than the baseline when reducing around $1G$ multiplications and additions.   

\begin{table}[t]
\caption{Experiment results on TinyImageNet.}
\label{tab:tinyimagenet_exp}
    \scriptsize
    \centering
    \setlength{\tabcolsep}{3mm}{
    \begin{tabular}{ccccc}
    \toprule 
    Model & Method & \#Add. & \#Mul. & Accuracy (\%) \\
    \midrule
    \multirow{3}{*}{Modified ConvMixer} & Baseline & $3.36G$ & $3.36G$ & $56.76$  \\
    ~ & PECAN-A & $2.36G$ & $2.36G$ & $59.42$ \\
    ~ & PECAN-D & $0.98G$ & $0$ & $50.48$ \\
    \bottomrule
\end{tabular}}
\end{table}

\section{Codes Instruction}
In order to make it easier to verify the experimental results, we provide codes and the running commands for readers to reproduce the results in the tables.

For dataset MNIST, CIFAR-10 and CIFAR-100, we train our models using the following commands.

\fbox{\shortstack[l]{
python train.py $\setminus$ $\quad \quad \quad \quad \quad \quad \quad \quad \quad \quad \quad \quad \quad \quad \quad \quad \quad \quad \quad \quad \quad \quad \quad \quad \quad \quad \quad \quad \quad \quad \quad$\\
--log\_dir [directory of the saved logs and models] $\setminus$\\
--data\_dir [directory to training data] $\setminus$\\
--dataset [MNIST/CIFAR10/CIFAR100]  $\setminus$\\
--arch [resnet20\_pecan\_a/resnet20\_pecan\_d] $\setminus$\\
--batch\_size [training batch] $\setminus$\\
--epochs [training epochs] $\setminus$\\
--learning\_rate [training learning rate] $\setminus$\\
--lr\_decay\_step [learning rate decay step] $\setminus$\\
--query\_metric [dot/adder] $\setminus$\\
--gpu [index of the GPU that will be used] 
}}

% \section{VGG and ResNet on CIFAR-100}

% In this section, we conduct experiments on CIFAR-100 adopting the same settings as shown in Appendix~\ref{appendix:p_setting_cifar10}. As can be seen, there still exist accuracy gaps for PECAN-D especially on ResNet20/32. One possible explanation is that more prototypes are required when applying PECAN to larger datasets, however, we keep the same settings as CIFAR-10 in this table. Additionally, we are dedicated to further improve the performance of PECAN-D on larger datasets including applying knowledge distillation, etc.

% \begin{table}[h]
% \centering
% \scriptsize
% \caption{Experiment results on CIFAR100.}
% \label{appendix:cifar100_exp}
% \vspace{1mm}
% \scriptsize
% \centering
% \setlength{\tabcolsep}{2.6mm}{
% \begin{tabular}{ccccc}
% \toprule 
% Model & Method & \#Add. & \#Mul. & Accuracy (\%) \\
% \midrule
% \multirow{3}{*}{VGG-Small} & Baseline & $0.61G$ & $0.61G$ & $67.84$  \\
% ~ & PECAN-A & $0.54G$ & $0.54G$ & $69.21$ \\
% ~ & PECAN-D & $0.37G$ & $0$ & $60.43$ \\
% \midrule
% \multirow{3}{*}{ResNet20} & Baseline & $40.56M$ & $40.56M$ &  $69.55$ \\
% ~ & PECAN-A & $38.12M$ & $38.12M$ & $63.15$  \\
% ~ & PECAN-D & $211.71M$ & $0$ &  $58.01$ \\
% \midrule
% \multirow{3}{*}{ResNet32} & Baseline & $68.86M$ & $68.86M$ & $70.57$  \\
% ~ & PECAN-A & $64.20M$ & $64.20M$ &  $64.13$ \\
% ~ & PECAN-D & $353.27M$ & $0$ &  $58.26$ \\
% \bottomrule
% \end{tabular}}
% \vspace{-3mm}
% \end{table}

\end{document}